%% file: KaRA-AI2ASE.tex
\documentclass[letterpaper]{article} 
\usepackage{aaai23}  
\usepackage{times}  
\usepackage{helvet}  
\usepackage{courier}  
\usepackage[hyphens]{url}  
\usepackage{graphicx} 
\usepackage{natbib}  
\usepackage{caption} 
\usepackage{amssymb}
\usepackage{amsmath}
\usepackage{booktabs}

\usepackage{tabularx}
\usepackage{rotating}
\usepackage{array}
\usepackage{subcaption}
\usepackage[table,xcdraw]{xcolor}

\usepackage{algorithm}
\usepackage{algorithmic}

\usepackage{newfloat}
\usepackage{listings}
\DeclareCaptionStyle{ruled}{labelfont=normalfont,labelsep=colon,strut=off} 
\floatstyle{ruled}
\newfloat{listing}{tb}{lst}{}
\floatname{listing}{Listing}
\title{Knowledge-augmented Risk Assessment (KaRA): a hybrid-intelligence framework for supporting knowledge-intensive risk assessment of prospect candidates}

\author {
    Carlos Raoni Mendes, 
    Emilio Vital Brazil, 
    Vinicius Segura, 
    Renato Cerqueira 
}
\affiliations {
    IBM Research\\
    Rio de Janeiro, RJ, Brazil \\
    craoni@br.ibm.com, evital@br.ibm.com, vboas@br.ibm.com, rcerq@br.ibm.com
}

\usepackage{bibentry}

\begin{document}

\maketitle

\input{tex/abstract}

\input{tex/introduction}

\input{tex/relatedwork}

\input{tex/KARAFramework}

\input{tex/prospectCharacterization}

\input{tex/LoK}

\input{tex/PoS}

\input{tex/conclusions}

\bibliography{references}

\end{document}

%% file: tex/abstract.tex
\begin{abstract}
Evaluating the potential of a prospective candidate is a common task in multiple decision-making processes in different industries. 
We refer to a prospect as something or someone that could potentially produce positive results in a given context, e.g., an area where an oil company could find oil, a compound that, when synthesized, results in a material with required properties, and so on. 
In many contexts, assessing the Probability of Success (PoS) of prospects heavily depends on experts' knowledge, often leading to biased and inconsistent assessments. 
We have developed the framework named KARA (Knowledge-augmented Risk Assessment) to address these issues. 
It combines multiple AI techniques that consider SMEs (Subject Matter Experts) feedback on top of a structured domain knowledge-base to support risk assessment processes of prospect candidates in knowledge-intensive contexts.
\end{abstract}

%% file: tex/introduction.tex
\section{Introduction}
\label{intro}
The assessment of the risk of failure or its opposite, the Probability of Success (POS), plays a crucial role in deciding which prospects are worth investing in. 
In many critical contexts, the prospect risk assessment process carries the characteristics of the so-called Knowledge-intensive process \cite{kip}. 
When this is the case, there is a strong dependency on the tacit knowledge of multiple experts in different areas, and a relevant part of the available data is heavily uncertain. Methods that don’t correctly handle this inherent complexity often lead to biased and inconsistent assessments. 
The challenge is to develop a methodology and supporting technology for prospect risk assessment in knowledge-intensive contexts that is consistent, reduces biases, controls uncertainty, and often leads to the selection of successful prospects (or discoveries). 
This work describes a framework that aims to address these issues. We called it the Knowledge-augmented Risk Assessment (KaRA) framework.

The recent advances in AI and Knowledge Engineering provided the basis for the development of KaRA. 
Knowledge engineering practices and technologies are applied to represent and integrate the domain knowledge from multiple data sources and stakeholders and provide easy access to this knowledge. 
At the same time, AI provides the appropriate tools for inference, prediction, uncertainty reduction, consensus reaching, etc. 
Then, KaRA combines multiple AI techniques that consider SME’s (Subject Matter Experts) feedback on top of a structured domain KB (Knowledge Base) to support the risk assessment processes of prospect candidates in knowledge-intensive contexts.  

The KaRA framework is a generalization and extension of the work we developed for supporting the assessment of the geological success of prospects (see \citet{gra-eage} and \citet{gsa-eage}). Next, we detail the general problem that KaRA approaches.

\subsection{Knowledge-intensive Prospect Risk Assessment}
\label{kipra}

In this work, we refer to a successful prospect as a discovery. Finding discoveries generally starts with many candidates from which very little is known. These candidates pass through a triage process that selects those worth further investigation. Given the number of candidates, the triage often involves some form of automatic filtering combined with experts’ simple evaluations. The candidates that pass triage became what we refer to as prospects. A prospect can pass through a detailed assessment that involves analyzing the many risk factors that could prevent it from succeeding. This process requires experts’ analysis from different areas and probably new data acquisition. These resource requirements imply the existence of a limited number of prospects. Then, a final investment is made to an even smaller number to become discoveries or failures. For instance, in oil and gas exploration, this last investment would mean drilling a well, while in materials discovery, that would mean synthesizing a new compound. Both examples are very resource-demanding, but they could generate high rewards on the other side. Therefore, the prospect selection process is essential to creating a successful business. At each step of the process the number of candidates decreases, while at the same time, more is known about them. Another essential aspect is that each step requires increasing resource investments.
\begin{figure*}[ht]
	\centering
	\includegraphics[width=0.75\textwidth]{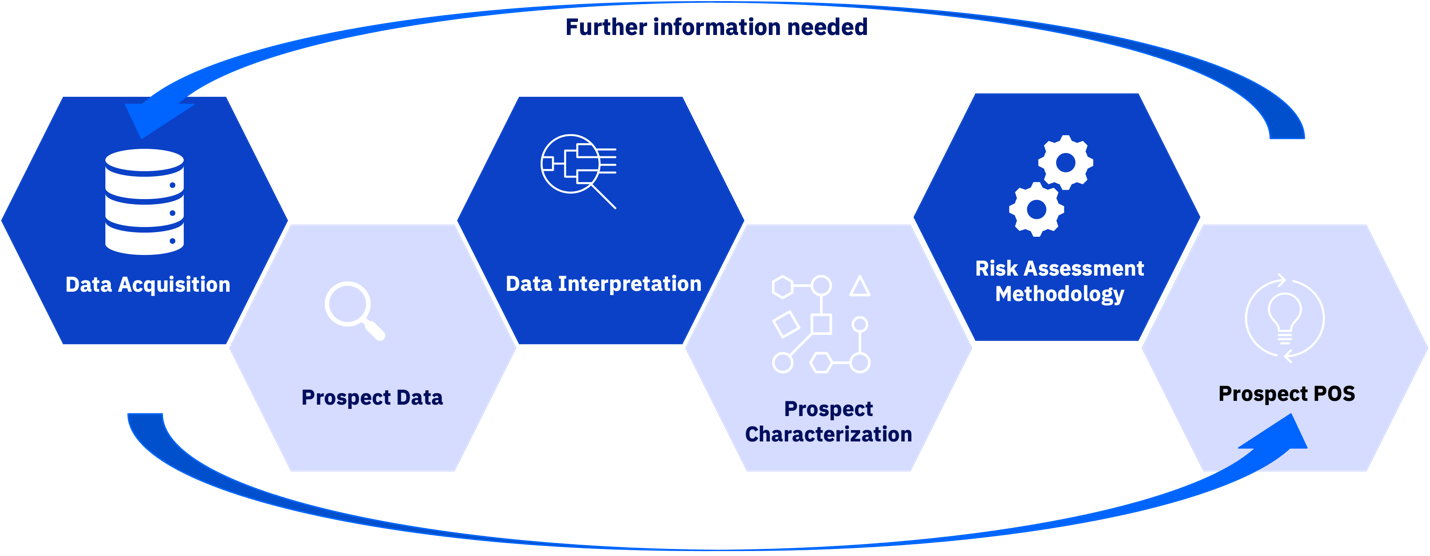}
	\caption{Prospect Risk Assessment Workflow}
	\label{fig-risk-assess-workflow}
\end{figure*}
In Figure \ref{fig-risk-assess-workflow}, we detail the workflow for assessing prospects’ POS. It starts with acquiring new data related to the prospect to be analyzed. The available data is then processed and interpreted by experts from different areas. This process is carried out to detail relevant information and characteristics resulting in a prospect characterization suited for the risk assessment methodology. It’s worth noting that the available data is often limited and uncertain. The experts’ interpretation compensates for these limitations with their experience and tacit knowledge. But the final characterization is still subject to uncertainty and may be biased by the experts’ opinions. The risk assessment methodology then takes the characterization as input and produces the prospect’s POS itself. It is also a process that depends on experts’ opinions and knowledge. It becomes only a guessing exercise for experts if done in an unstructured form, often leading to inconsistent and biased assessments \cite{milkov2015risk}.

The prospect’s POS helps decision-makers decide if they should invest, discard, or need more information/data and then another round of assessment. This last decision is represented in Figure \ref{fig-risk-assess-workflow} by the backward arrow. Essentially an enduring decision (invest or discard) is made when there is enough confidence in the current knowledge of the prospect and, consequently, in the assessment. Some risk assessment methodologies use the metric Level-of-Knowledge (LOK) to represent how much is known about the prospect explicitly \cite{lowry2005advances}.

%

\subsection{Structure}

In the next section, we detail the related literature with a focus on the works in the Oil \& Gas exploration space. 
We assume that it is the most advanced industry regarding prospect risk assessment. 
In the section \textit{KaRA Framework}, we present an overview of the proposed methodology developed in KaRA. 
Subsequent sections are dedicated to detail the methodology phases, that are, prospect characterization, Level-of-Knowledge assessment, and Probability-of-Success assessment. 

%% file: tex/relatedwork.tex
\section{Related Literature}
\label{rel-literature}

The generalization that we proposed for knowledge-intensive prospect risk assessment was derived from the Oil \& Gas exploration domain. In this context, a prospect is a region containing a potentially recoverable oil accumulation. To confirm or discard the hypothesis of oil accumulation in the prospect region, the company must drill a well, which is very resource demanding. It is a high-risk, high-reward situation, so this industry treats it very carefully. We believe that prospect risk assessment in the Oil \& Gas exploration space is mature compared to other prospect exploration domains. Therefore, we will use its literature as a reference in this section.

Given the complexity involved in assessing a prospect, a common approach adopted by the industry is to break the assessment into different geological (or risk) factors, evaluating the POS of each one and then combining the corresponding probabilities to obtain the POS for the prospect itself. The factors correspond to geological features that should exist to form an oil accumulation. One common assumption to facilitate the analysis is to adopt independence between the risk factors. In this case, the prospect’s POS is the multiplication of all individual factor’s POS values. So, choosing how to break the assessment into different factors plays a crucial role in the methodology. An excessive number may result in very small and hard to differentiate prospects’ POS values, while a small number may neglect essential dimensions of the risk evaluation. So, the methods adopted across different organizations vary in the number of geological risk factors analyzed. \citet{milkov2015risk} presents a table comparing various methodologies found in the literature. To illustrate this variation, the method proposed by \citet{lowry2005advances} considers the assessment of 19 geological risk factors, while \citet{rose2001risk} analyzes only five.

\begin{figure*}
	\centering
	\includegraphics[width=0.8\textwidth]{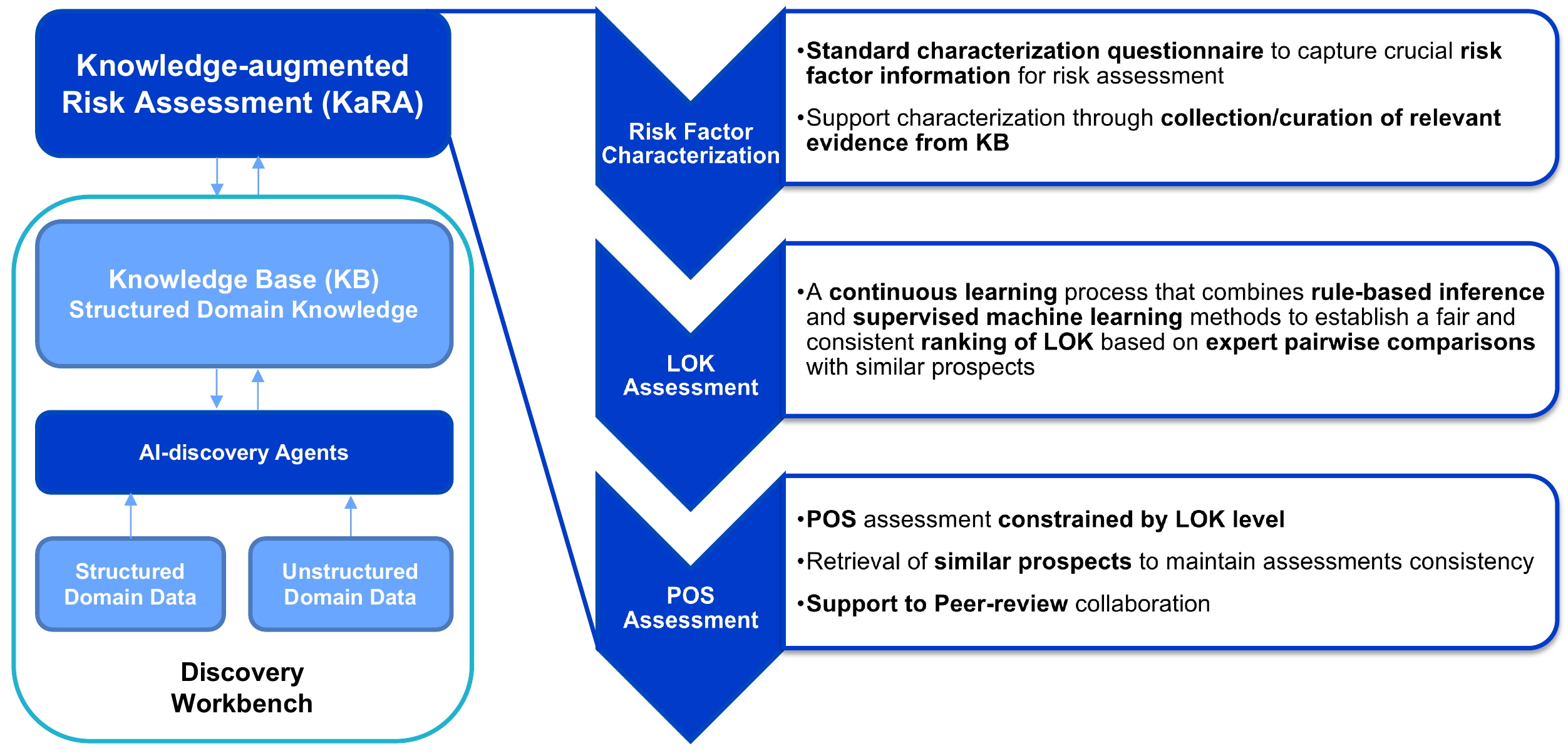}
	\caption{Overview of KaRA – methodology workflow and features.}
	\label{fig-kara-overview}
\end{figure*}

The methodologies also vary in the strategy that guides POS assessments for each geological factor. They must balance the quality and quantity of the available data with the geological model characteristics to fine-tune the range of possible values experts could adopt for the POS. In \cite{milkov2015risk} and \cite{jan2000ccop}, the authors propose using reference tables. To use a geological factor reference table, one must identify which cell the prospect maps based on the available data and geological model characteristics. While in \cite{milkov2015risk}, the cell specifies the POS value itself, in \cite{jan2000ccop}, they provide a range of possible values that POS could assume. Although adopting these tables may improve the consistency and repeatability of the assessment process, they are not suitable to accommodate new disruptive knowledge. One clear example of this problem is that the tables become outdated when new technologies are developed.

Another popular method is the use of the metric Level-of-Knowledge (LOK) to assess data availability, quantity, and quality (see \citep{paltrinieri2019learning}). Some approaches use the LOK metric to constrain the possible range of values assigned to the POS. In \cite{rose2001risk}, we find a matrix specifying a relation where the POS is restricted to the middle of the scale (around fifty percent) when the LOK is low. At the same time, at high LOK levels, the POS value should be either very low (around zero percent) or very high (around a hundred percent). The rationale is that in a low confidence situation, one should not be conclusive about the POS value, while in high confidence cases, one must be very assertive. In \cite{lowry2005advances}, they apply the same rationale and create a graphic that specifies a LOK and POS relationship by delimiting a region of allowed LOK and POS assessment pairs. 
The same work also presents alternative LOK/POS relationships that implement different rationales. The authors also recognize that making quantitative assessments of LOK and using it to manipulate POS is challenging both in theory and practice.

The KaRA framework aims at addressing the limitations of the described methods. The main objective is to provide a methodology that guarantees consistency, decreases biases, and continuously improves given new relevant knowledge obtained from data sources and or SMEs, all supported by practical and meaningful LOK quantitative scales.

%% file: tex/KARAFramework.tex
\section{KaRA Framework}
\label{methodology}

The KaRA framework is a generalization and extension of the work we developed for supporting the assessment of the geological success of prospects (see \citet{gra-eage}, and \citet{gsa-eage}). It combines multiple AI techniques that consider SMEs' feedback on top of a structured domain knowledge base to support the risk assessment processes of candidates in knowledge-intensive contexts. The recent advances in AI and Knowledge Engineering provided the basis for the development of KaRA. Knowledge engineering practices and technologies are applied to represent and integrate the domain knowledge from multiple data sources and stakeholders and provide easy access to this knowledge. At the same time, AI provides the appropriate tools for inference, prediction, uncertainty reduction, consensus reaching, etc. 

The methodology implemented by KaRA divides the assessment into multiple risk factors. Individual risk factor evaluation encompasses three phases: risk factor characterization, LOK assessment, and POS assessment. Different AI agents support expert decisions at each one of them. As mentioned, the domain knowledge is structured and integrated by the KB. The KB also handles all the data used by KaRA components. KaRA aims to enable a co-creation environment for SMEs and AI agents to support the risk assessment process. Figure \ref{fig-kara-overview} summarizes the architecture, the methodology workflow, and some of the main features of KaRA. We discuss those features in the sequence according to each phase.

\begin{figure*}[ht]
	\centering
	\includegraphics[width=0.85\textwidth]{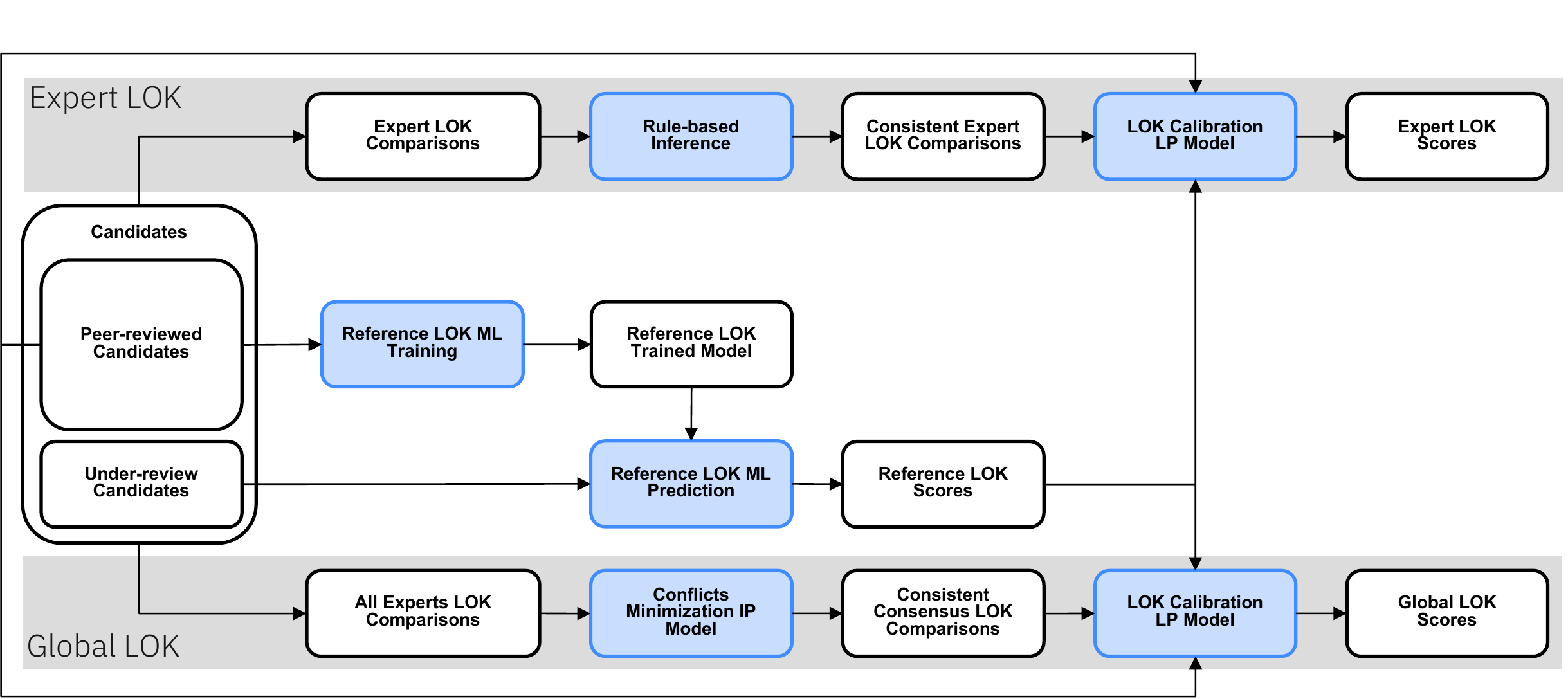}
	\caption{Workflows for expert and global LOK scores.}
	\label{fig-lok-workflows}
\end{figure*}

%% file: tex/prospectCharacterization.tex
\subsection{Risk Factor Characterization} \label{kara-characterization}

We adopted a similar approach to \cite{milkov2015risk} and \cite{jan2000ccop} in the risk factor characterization phase. For each risk factor, experts should answer a standard questionnaire showcasing the data analysis workflow results that could influence LOK or POS assessments. The characterization questionnaires structure the candidates in the KB, providing an easy way to represent them for supervised and unsupervised machine learning algorithms. One representative example used in the following phases is retrieving similar candidates in terms of characterization for comparison purposes. The differentiator of KaRA will be the ability to query and reason through the KB to retrieve and rank relevant evidence that may help the experts answer the characterization questions. The expert will curate the evidence recovered from the KB, keeping only the ones relevant to the answer. This feedback feeds the ranking algorithms used in the evidence retrieval, which are continuously improved. This mechanism establishes a co-creation process where AI and experts collaborate to accurately characterize the candidate risk factors taking advantage of available evidence stored in the KB and the expert's tacit knowledge.


%% file: tex/LoK.tex
\subsection{LOK Assessment} \label{kara-lok-assessment}

The role of the LOK assessment phase is to establish fair and consistent LOK score scales dependent on the risk factor characterizations produced in the previous step to support the POS assessment process, such as done in \cite{lowry2005advances} and \cite{rose2001risk}. The LOK scales are created based on pairwise comparisons made by the experts and previously assessed, and peer-reviewed candidates present in KB. These candidates serve as training examples for a supervised machine learning algorithm that gives an initial LOK estimate, the so-called reference LOK. Experts' pairwise LOK comparisons calibrate this estimate by solving a Linear Programming (LP) model that determines the smallest overall adjustments in the LOK values needed to respect the comparisons. A rule-based inference strategy maintains the consistency of comparisons for a particular individual. The comparisons at the individual level capture the ranking opinions of each expert. The reference LOK estimates calibrated by these comparisons create the expert LOK scale. The system also determines the subset of consistent LOK comparisons that best represent experts' consensus by solving an Integer Programming (IP) model that minimizes the conflicts between the output comparisons and the overall expert opinions. We obtain the so-called global LOK scale when these comparisons calibrate the reference LOK estimates with the same LP model previously described. Additional candidates and comparisons continuously adjust expert and global LOK scales. Figure \ref{fig-lok-workflows} summarizes expert and global LOK workflows. This approach is another example of a human-AI co-creation strategy applied to support an essential step in the discovery process.  

\subsubsection{LOK comparisons consistency trough rule-based inference}

\begin{figure}[ht]
    \centering
    \includegraphics[width=.3\linewidth]{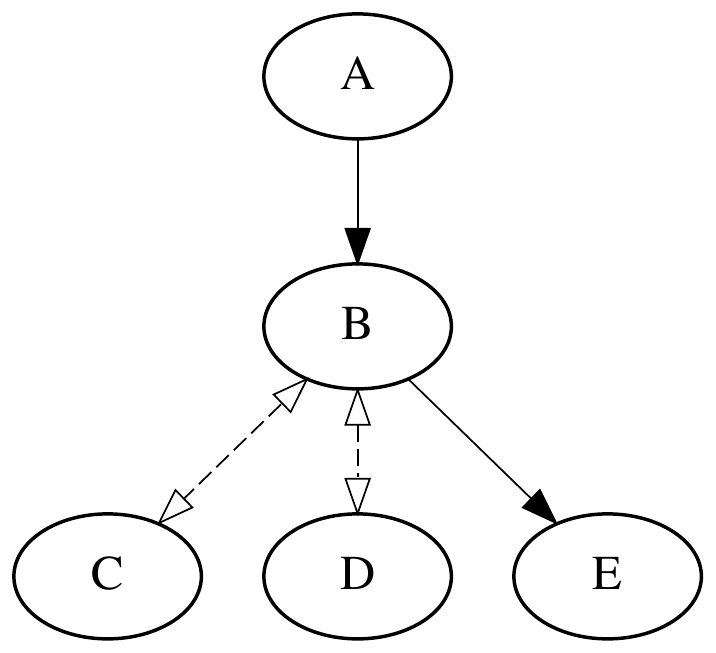}
    \caption{Canonical LOK comparisons graph}
    \label{fig:expert-canonical-lok-graph}
\end{figure}

\begin{figure}[ht]
	\centering
	\includegraphics[width=.3\linewidth]{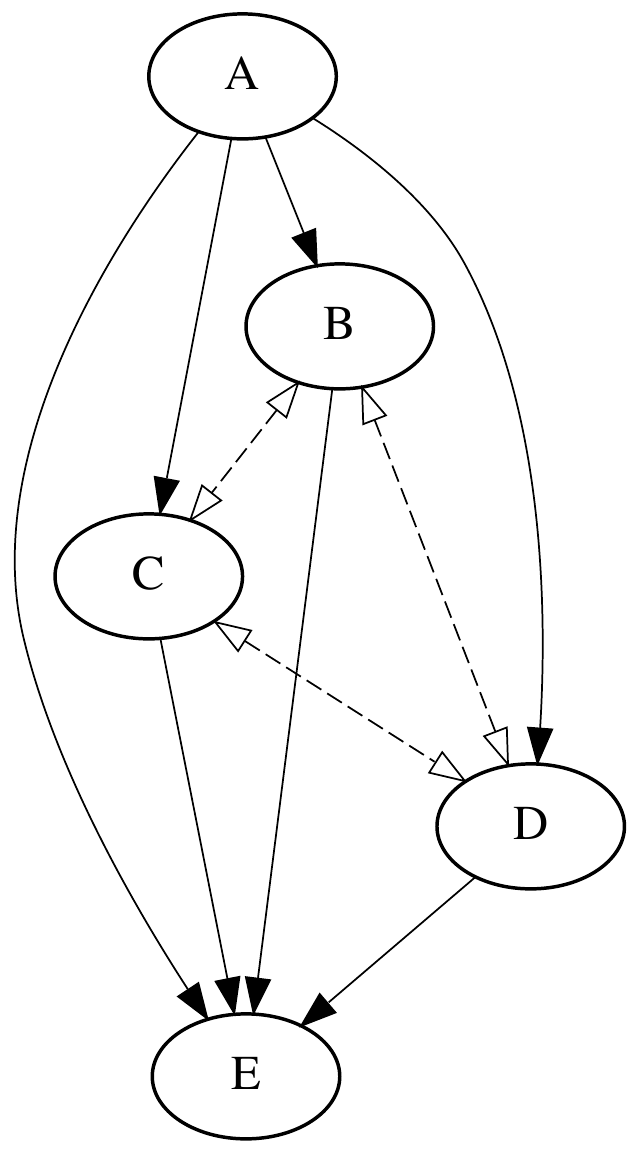}
	\caption{Consistent LOK comparisons graph after inference}
	\label{fig:expert-inferred-lok-graph}
\end{figure}

As previously mentioned, when evaluating a risk factor of a target prospect, experts compare pairwisely the risk factor characterization of the target prospect with other similar risk characterizations present in the KB. In the pairwise comparison, the expert answers which risk characterization should receive a higher LOK score or if their LOK scores should be the same. A rule-based approach keeps the consistency of LOK comparisons for each expert. Given risk factor characterizations $a$, $b$, and $c$, and $L_i$ representing the LOK score of characterization $i$, the consistency rules for pairwise LOK comparisons are: 
\begin{itemize}
    \item $(L_a < L_b) \land (L_b < L_c) \implies L_a < L_c$
    \item $(L_a = L_b) \land (L_b = L_c) \implies L_a = L_c$
    \item $(L_a < L_b) \land (L_b = L_c) \implies L_a < L_c$
\end{itemize}
In figure \ref{fig:expert-canonical-lok-graph}, we show a graph representation of canonical expert's LOK comparisons. The nodes represent the risk factor characterizations, directed edges from a node $i$ to a node $j$ represent a LOK comparison where $L_i > L_j$, while the bi-directional ones represent a LOK comparison $L_i = L_j$. 
Figure \ref{fig:expert-inferred-lok-graph} shows the graph representation after applying LOK consistency rules. The inference of those rules could be obtained by polynomial-time graph-based algorithms such as the ones found in \cite{mohr1986arc}.

\subsubsection{Reference LOK and LOK Calibration LP}
\label{lok-calibration}
Reference LOK scores provide a first quantitative estimation of LOK based on peer-reviewed examples, which means that a group of experts validated their final LOK assessment scores. Using a one-hot encoding representation of risk-factor characterizations, we apply supervised machine learning (ML) algorithms to estimate the reference LOK scores. The model is re-trained whenever a new peer-reviewed assessment is stored in the KB. We use the average loss of a ten-fold cross-validation test to choose the best ML model (i.e., algorithm and parametrization). Despite the usefulness of reference LOK, we adopt in KaRA an approach that prioritizes expert feedback. We create individual LOK scales by calibrating the reference LOK estimates with the expert's consistent LOK comparisons. Given a set of risk factor characterizations $P$, reference LOK estimates $L_{Ri}$ ($ \forall i \in P$), a set of consistent greater than $GT$ and equals $EQ$ pairwise $(i,j)$ LOK comparisons, the following LP is used to estimate the expert's LOK scores $L_{i}$ ($\forall i \in P$): 
\begin{align} 
	{\textrm{Min}} \quad \sum_{i \in P}| L_{i} - L_{Ri} |  & \label{calib-obj}\\
	s.t. \hspace{0.3\linewidth} & \nonumber \\
	L_{i} - L_{j} \geq t  \quad & \forall (i,j) \in GT \label{calib-gt}\\
	L_{i} - L_{j} = 0 \quad & \forall (i,j) \in EQ \label{calib-sim}\\ 
	0 \leq L_{i} \leq 1 \quad & \forall i \in P \label{calib-domain}
\end{align}
The objective function (\ref{calib-obj}) minimizes the absolute difference between the expert's LOK scores $L_i$ and the reference LOK $L_{Ri}$. Constraints (\ref{calib-gt}) guarantee that the LOK score of $i$ is greater than the LOK score of $j$ by at least a threshold $t$ for each greater than comparison $(i, j) \in GT$. Similarly, constraints (\ref{calib-sim}) guarantee that equality pairwise comparisons $(i,j) \in EQ$ are respected. Constraints (\ref{calib-domain}) define the domain of LOK scores to be between $0$ and $1$. Despite the absolute value function (\ref{calib-obj}), we can transform this formulation into a standard form LP following strategies such as described in \cite{bertsimas1997introduction}.

\subsubsection{Global LOK Conflicts Minimization (or Consensus) IP}
\label{global-lok-model}
As previously explained, the framework maintains a set of consistent LOK comparisons for each expert. So, the opinions of different experts may conflict regarding the LOK pairwise relation for pair of characterizations $a$ and $b$, which could be higher ($L_a > L_b$), lower ($L_a < L_b$) or equal ($L_a = L_b$). In this context, achieving overall consistency would mean discovering the set of pairwise LOK relations that minimizes the conflicts with all expert opinions. In \cite{brancotte2015rank}, authors proposed an Integer Programming (IP) model to solve a ranking aggregation problem with ties and partial orderings. We used this formulation as a reference to create a new one with fewer constraints and suitable for our context. Next, we describe each element used in the formulation and present it in the sequence: 
\begin{itemize}
	\item $x_{a\leq b}$ -- binary decision variable that indicates if $L_a \leq L_b$.
	\item $x_{a = b}$ -- binary decision variable that indicates if $L_a = L_b$.
	\item $w_{a\leq b}$ -- constant representing the number of experts that chose lower than or equal as pairwise LOK relation between $a$ and $b$.
	\item $w_{a = b}$ -- constant representing the number of experts that chose equal as pairwise LOK relation between $a$ and $b$.
\end{itemize}
\begin{align} 
	{\textrm{Min}} \quad \sum_{a} \sum_{b} (w_{a \leq b}*x_{b\leq a} + \hspace{0.1\linewidth} &\nonumber \\
	w_{b \leq a}*x_{a \leq b} - 2(&w_{a=b}) * x_{a=b})  \label{objective_global_lok} \\
	s.t. \hspace{0.55\linewidth} &  \nonumber \\	
	- x_{a\leq b} - x_{b\leq a} \leq -1 \ \ \forall a \in P&, b \in P \label{r1_raoni_v1}\\
	2x_{a=b} - x_{a\leq b} - x_{b\leq a} \leq 0\  \ \ \forall a \in P&, b \in P \label{r2_raoni_v1} \\
	x_{a\leq b} + x_{b\leq a} - x_{a=b} \leq 1\ \ \ \forall a \in P&, b \in P \label{r3_raoni_v1} \\
	x_{a\leq c} + x_{c\leq b} - x_{a\leq b} \leq 1\ \ \ \forall a \in P&, b \in P, c \in P \label{r4_raoni_v1} \\
	x_{a\leq b} \in \{0, 1\}\ \ \ \forall a \in P&, b \in P \\
	x_{a=b} \in \{0, 1\}\ \ \ \forall a \in P&, b \in P 
\end{align}
The rationale of this formulation is to guarantee that $x_{a \leq b} = 1$ and $x_{b \leq a} = 1$ implies $x_{a = b} = 1$. This way, a strict lower relation happens when $x_{a \leq b} = 1$ and $x_{a = b} = 0$. The objective function (\ref{objective_global_lok}) considers this assumption to represent the minimization of conflicts.
The constraints (\ref{r1_raoni_v1}) ensures that at least one relation is adopted for each pair of elements $a$ and $b$.
The constraints (\ref{r2_raoni_v1}) and (\ref{r3_raoni_v1}) guarantees that, if both $x_{a\leq b}$ and $x_{b\leq a}$ are one, than $x_{a=b}$ will be as well. Finally, constraint (\ref{r4_raoni_v1}) ensures that transitivity is maintained throughout the solution.



%% file: tex/PoS.tex
\subsection{POS Assessment} \label{kara-pos-assessment}
\begin{figure*}
    \centering
    \begin{subfigure}{0.49\linewidth}
        \includegraphics[width=\linewidth]{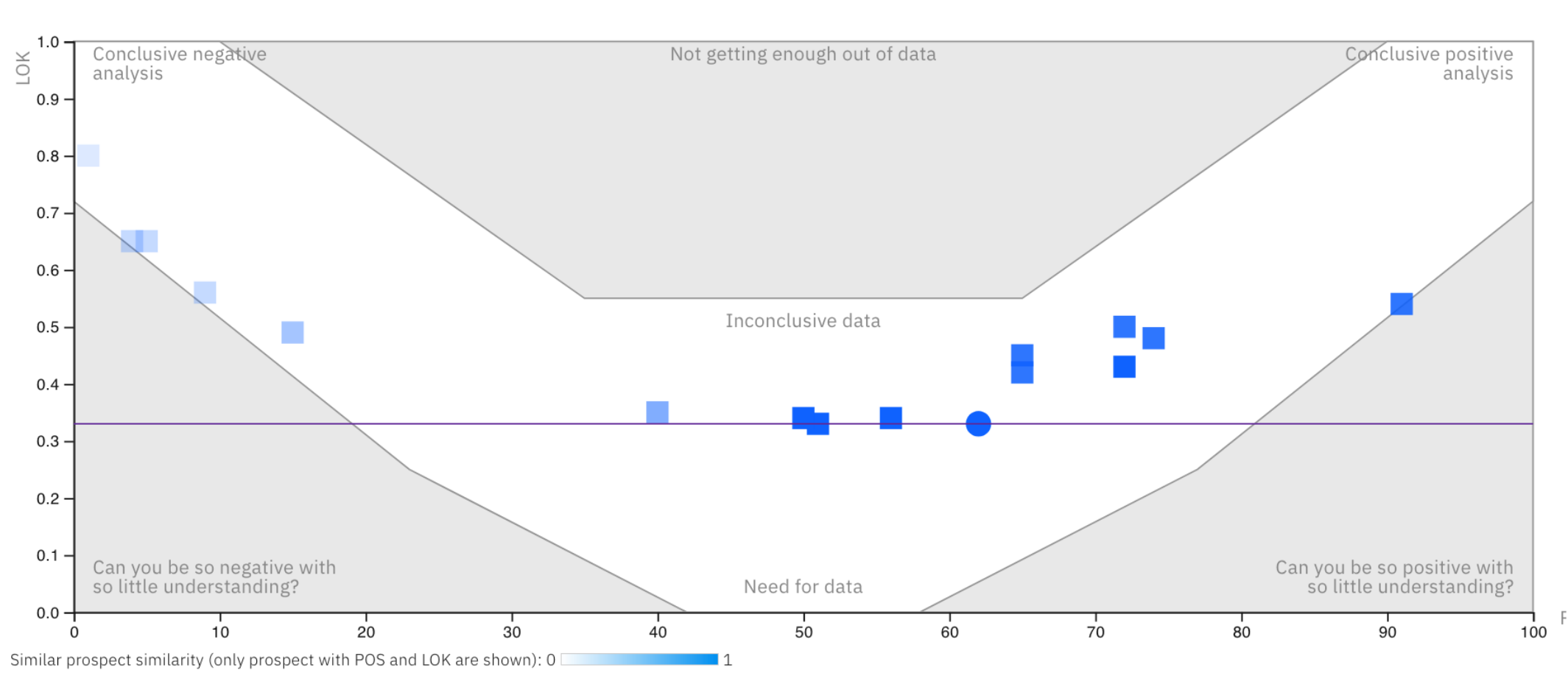} 
        \caption{Expert POS assessment}
        \label{fig:expert-pos-assessment}
    \end{subfigure}\hfill
    \begin{subfigure}{0.49\linewidth}
        \includegraphics[width=\linewidth]{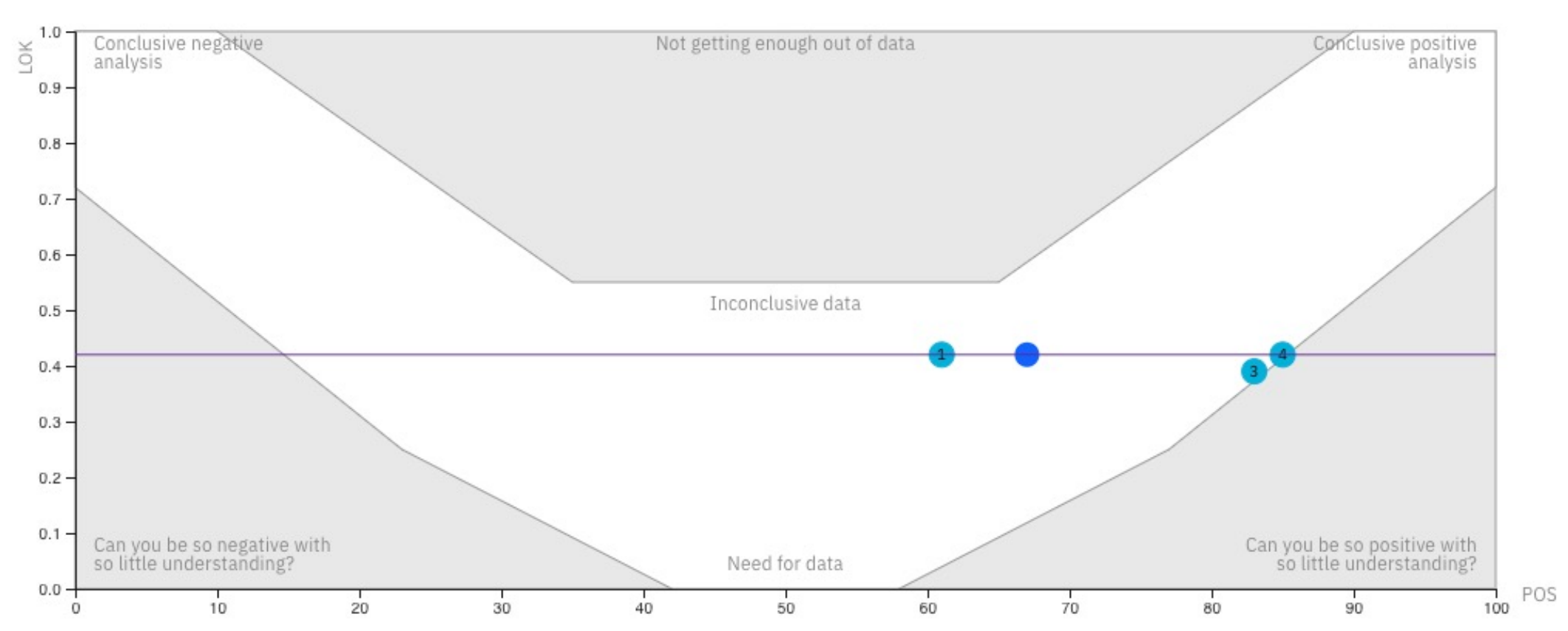}
            \caption{Peer-review POS assessment}
        \label{fig:peer-review-pos-assessment}
    \end{subfigure}
    \caption{Confidence-likelihood plots.}
    \label{fig:hornplots}
\end{figure*}

In the last phase, experts' should collaborate to give a final POS value to the risk factor of the target candidate. The process collects individual POS assessments of multiple experts, and then a peer-review evaluation is conducted to reach a consensus value. Both steps constrain the experts' possible POS assessment values given a certain LOK level. The specification of how the LOK constrains the POS values is part of the specialization of the framework. KaRA's default method uses the so-called confidence-likelihood plot, a strategy devised in the oil and gas industry (see \citet{rose2001risk} and \citet{lowry2005advances}). In this approach, POS values are constrained to the middle of the scale (around fifty percent) when the LOK is low, while at high LOK levels, the POS value should be either very low (around zero percent) or very high (around a hundred percent). The rationale is that in a low confidence situation, one should not be conclusive about the POS value, while in high confidence cases, one must be very assertive. The plots shown in Figure \ref{fig:hornplots} specify this relationship by delimiting a region (in white) of allowed LOK (y-axis) and POS (x-axis) assessment pairs. 

Figure \ref{fig:expert-pos-assessment} shows the supporting plot for the individual expert's POS assessment step, while Figure \ref{fig:peer-review-pos-assessment} presents the one for the peer-review POS assessment. In Figure \ref{fig:expert-pos-assessment}, we use the expert's LOK score to position the line on the y-axis. The expert is then constrained to give a POS value in the intersection between this line and the white region. The blue squares represent the assessments of similar candidates, where the blueness is proportional to the similarity with the target candidate. They serve as a reference to possibly maintain consistency with prior assessments. In Figure \ref{fig:peer-review-pos-assessment}, the global LOK score determines the position of the LOK line, and the circles represent the multiple assessments of the different experts. With both information, the peer-review step may reach a final consensus driven by the multiple experts' opinions, potentially decreasing the biases in the process.

%% file: tex/conclusions.tex
\section{Conclusions and Future Work}
\label{conclusions}
We described a knowledge-intensive prospect risk assessment that can be applied to critical business decisions in multiple domains. 
To the best of our knowledge, KaRA is the first framework to address this problem in a general sense. 
The methodology proposed in the KaRA framework combines AI and knowledge engineering to address the limitations of the methods found in the literature. 
Guarantee consistency, decreasing biases, capturing experts' knowledge, being flexible to adapt to technology disruptions, and providing a practical and meaningful LOK quantitative scale are some of the most relevant objectives of the research around KaRA.
It may become essential and provide a competitive edge in multiple accelerated discovery pipelines.
We have tested KaRA in two domains, petroleum exploration, and material discovery, demonstrating the flexibility of our approach. 
Moreover, We have feedback from many experts in both areas about the positive impact of our system.
However, unfortunately, we do not have quantity results to be discussed at this point. 
Even though we have experts using KaRA nowadays, capturing data to analyze the system's effectiveness is problematic because it involves knowledge-intensive tasks.
To approach this issue, we plan to organize studies to measure the effects of the KaRA using controlled data and focus groups.
We also will develop a link between the process of creating the methodology (the questionary) and the knowledge base. 
There is a synergy between creating the questions and feeding an ontology that can enrich the system, enabling reasoning about data already captured to support answering some questions.